# Empirical Strategy for Stretching Probability Distribution in Neural-network-based Regression

Eunho Koo[1,2] and Hyungjun Kim[3]

[1]*Center for Mathematical Analysis and Computation, Yonsei University, Seoul, South Korea*

[2]*LTS, Inc., Tokyo, Japan*

[3]*Institute of Industrial Science, The University of Tokyo, Tokyo, Japan*

September 3, 2020

\* Corresponding author: Hyungjun Kim

Institute of Industrial Science, The University of Tokyo, Tokyo, Japan

Email: hjkim@iis.u-tokyo.ac.jp

Abstract

In regression analysis under artificial neural networks, the prediction performance depends on determining the appropriate weights between layers. As randomly initialized weights are updated during back-propagation using the gradient descent procedure under a given loss function, the loss function structure can affect the performance significantly. In this study, we considered the distribution error, i.e., the inconsistency of two distributions (those of the predicted values and label), as the prediction error, and proposed weighted empirical stretching (WES) as a novel loss function to increase the overlap area of the two distributions. The function depends on the distribution of a given label, thus, it is applicable to any distribution shape. Moreover, it contains a scaling hyperparameter ($\beta$) such that the appropriate parameter value maximizes the common section of the two distributions. To test the function capability, we generated ideal distributed curves (unimodal, skewed unimodal, bimodal, and skewed bimodal) as the labels, and used the Fourier-extracted input data from the curves under a feedforward neural network. In general, WES outperformed loss functions in wide use, and the performance was robust to the various noise levels. The improved results in RMSE for the extreme domain (i.e., both tail regions of the distribution) are expected to be utilized for prediction of abnormal events in non-linear complex systems such as natural disaster and financial crisis.

1. INTRODUCTION

Numerous studies have been conducted that have focused on methodologies for artificially changing the distribution of each node and weight in a given architecture to obtain improved results, particularly in terms of the activation functions (Maas, Hannun, & Ng, 2013; Leshno, Lin, Pinkus, & Schocken, 1993; Glorot, Bordes, & Bengio, 2011), initialization of weights (Glorot & Bengio, 2010), and loss functions (Falas & Stafylopatis, 1999). In particular, the loss function in supervised learning provides mathematical criteria to estimate the loss of the model and thereby updates weights to reduce the loss at each training epoch; therefore, it plays a central role in artificial neural networks (Falas & Stafylopatis, 1999).

In general, loss functions can be categorized into two groups: classification and regression. In the case of classification, the cross-entropy function (Kapur & Kesavan, 1992) has been reported as an appropriate function; the definition of the function stems from information theory, and the function has been used for measuring the difference between two probability distributions (Murphy, 2012). A supplementation of the cross-entropy function could be the deformation of the log-likelihood loss function (Kanamori, 2010) (for multiclass applications), which creates a clear correspondence between the decision functions and conditional probabilities of the label. Another loss function for classification is the hinge loss function (Rosasco, De Vito, Caponnetto, Piana, & Verri, 2004), which is applicable to the support vector machine (Cortes & Vapnik, 1995). In hinge loss, if the signs of the values of the label and prediction are opposite, significant penalties are applied, thereby inducing the values to have the same sign. The focal loss function (Lin, Goyal, Girshick, He, & Dollar, 2017), which is an enhanced version of the cross-entropy function, has been employed in unbalanced classification situations. It assigns different weights to the loss according to the corresponding frequencies of each class.

In regression, in addition to the mean square error (MSE) function, which has been the most common loss function, the Huber loss function (Huber, 1964) has been used for robust regression. The function was initially developed to penalize errors induced by an estimation procedure so that the performance would not be critically affected by outliers. Moreover, the reinforced pseudo-Huber loss function (Charbonnier, Blanc-Féraud, Aubert, & Barlaud, 1997; Hartley, Zisserman, Richard, Hartley, & Zisserman, 2003) has been reported as substitutable for application in deep neural networks, because its derivatives are continuous for all degrees. In statistics and econometrics, the quantile loss function (Koenker, 2015) has been used for estimating either the median or other quantiles of outputs, whereas MSE is applied for the resulting mean of outputs. In addition to its robustness against outliers in measurements, providing different measures of central tendency and statistical dispersion, the function may be useful in analyzing relationships among variables. The log hyperbolic cosine (*lncosh*) loss function has been suggested to control the input and output data when the observations are subject to unknown noise distributions (Karal, 2017). The function is also used for maximum likelihood parameter estimation in the construction of the optimum synchronizer (Wintz & Luecke, 1969).

In this study, we propose weighted empirical stretching (WES) as a novel loss function, which is dependent on the probability distribution of the label and is robust against noisy input data in regression. To illustrate our motivation, consider a regression experiment with a Gaussian distributed label. Under certain selections of the weight initialization method and activation function with the existing loss function, the predicted values generally exhibit a narrower Gaussian distribution than that of the label. This phenomenon occurs because most values in the label are concentrated near the mean and the loss function attempts to reduce only the mean value of the total errors. This phenomenon exhibits certain drawbacks, especially in the prediction of the extreme values corresponding to both tail parts in the distribution graph. Also, the prediction performance suffers from noisy data originating from factors such as measurement errors and human labeling errors. Typical approaches to overcome this problem have focused on data-cleaning methods and generative models. The former involves identifying corrupted data points (Zhu, Wu, & Chen, 2003; Brodley & Friedl, 1999; Angelova, Abu-Mostafa, & Perona, 2005), whereas the latter treats the unknown true labels as hidden variables and estimates a generative model (Reed et al., 2015; Xiao, Xia, Yang, Huang, & Wang, 2015; Bootkrajang & Kabán, 2012). To overcome the abovementioned challenges, we suggest WES as a method for increasing the area of the overlapping section between the label and prediction distributions, which results in the robust performance against noisy input data. To demonstrate the performance and applicability of WES, we employ four distribution types of manipulated ideal data: unimodal, skewed unimodal, bimodal, and skewed bimodal. Unimodal distribution represented by the Gaussian distribution has been the most frequently observed distribution. Skewed unimodal distribution often occur, such as the distribution of household income (Kennedy, Kawachi, Glass, & Prothrow-Stith, 1998) and the time intervals between coal mining disasters (Jarrett, 1979). Bimodal distribution can be observed in phenomena such as financial activities in social system (Scalas et al., 2004) and tsunami in nature (Geist & Parsons, 2008). Skewed bimodal distribution is easily found in complex natural system such as East Asia Summer Monsoon (Wang & Ho, 2002).

The remainder of this paper is outlined as follows. Section 2 summarizes the basic feedforward network (FFN) algorithm and defines the concrete structure of the WES loss function. Section 3 introduces methods of manipulation of various ideal data as the label and the extraction of Fourier signals from the curve as input data. The experimental setup and results are presented in Section 4, and we conclude the paper in Section 5.

## 2. METHODS

A feedforward neural networks (Elman, 1990; Hornik, Stinchcombe, & White, 1989) is the first and simplest type of neural network (Schmidhuber, 2015). In this architecture, the information propagates from the input node, via the hidden nodes, to the output node. The FFN uses various learning techniques involving back-propagation, enabling it to determine the appropriate weights located between layers to minimize the difference between the predicted values and label. Following a sufficiently large number of training epochs, the architecture converges to a certain state of optimization, and the network is referred to as learned.

We denote the total number of layers as $L$ and the number of nodes for each layer as $n_i$ for all $i = 1, 2, \ldots, L$, with $n_L = 1$. Moreover, we denote the loss and activation functions as $J$ and $\sigma$, respectively. The overall procedures can be divided into two parts, namely feedforwarding and back-propagating.

Feedforwarding:

$$z_{i+1} = \Theta_i a_i + b_i \quad (1)$$
$$a_{i+1} = \sigma(z_{i+1}), \quad (2)$$

where $a_i \in \mathbb{R}^{n_i}$, $\Theta_i \in \mathbb{R}^{n_{i+1} \times (n_i+1)}$, and $b_i \in \mathbb{R}$ correspond to vectors in the layers, weight matrix, and bias value, respectively, for $i = 1, 2, \ldots, L-1$.

Back-propagating:

$$\delta_L = \frac{dJ}{da_L} * \sigma'(z_L), \quad \delta_i = (\Theta_i^T \cdot \frac{dJ}{da_L}) * \sigma'(z_L) \qquad i = L-1, L-2, \ldots, 2 \quad (3)$$

$$\frac{\partial J}{\partial \Theta_i} = \delta_{i+1} \cdot a_i^T, \quad \frac{\partial J}{\partial b_i} = \delta_{i+1} \qquad i = 1, 2, \ldots, L-1 \quad (4)$$

$$\Theta_i \leftarrow \Theta_i - \alpha \frac{\partial J}{\partial \Theta_i}, \quad b_i \leftarrow b_i - \alpha \frac{\partial J}{\partial b_i} \qquad i = 1, 2, \ldots, L-1, \quad (5)$$

where $\Theta_i^T$ denotes the transpose of $\Theta_i$, $\sigma'(x)$ denotes the differentiation of $\sigma$ with respect to $x$, "$*$" denotes the element-wise product of two vectors, $\alpha$ denotes the learning rate, and "←" implies the updating of weights. All vectors are expressed in a column form, and "·" indicates the matrix multiplication operator.

The loss function $J$ serves as a crucial function in updating the weights during back-propagation. With a focus on matching the distributions of the predicted values and labels, we need to address the concept of imposing a small penalty on the label region with a higher probability density, and vice versa. Therefore, it is straightforward to define the loss function that depends on the probability density function (PDF) of the label. To implement such feature, we suggest the following WES as the loss function:

$$J(\hat{y}_i, y_i) = \frac{1}{N}\sum_{i=1}^{N}[\frac{1}{2}(\hat{y}_i - y_i)^2 \times g(y_i)], \tag{6}$$

where

$$g(x) = (\beta - c) \times \left(-\frac{1}{max_x(f(x))}f(x) + 1\right) + c. \tag{7}$$

In this case, $f(x)$ is the probability density function of the label, whereas $\hat{y}_i$ and $y_i$ are the predicted value and label corresponding to the $i$th input data, respectively. Moreover, $c$ and $\beta$ are values to be determined. The number $c$ is related to the range of a given label; if the range of the label is $[y_{min}, y_{max}]$, $c = \frac{1}{y_{max}-y_{min}}$. In this experiment, because the uniform normalization is used, which transforms to have the maximum and minimum values of the label as 1 and 0, respectively, the label range is $[0, 1]$, and hence, $c$ becomes 1. $\beta$ varies so that the loss function on a certain region of the distribution is penalized according to the probability density of that region (larger value of $\beta$ results in a higher penalty on the region with a lower frequency). There are five further remarks on WES.

- If the label is uniformly distributed, according to (7), $max_x(f(x)) = f(x) = Const$. Hence, $g(x) = c$ for all $x$ in its domain; that is, the loss function $J$ in (6) is simply the MSE loss function, which is not an interesting case.

- For the $i$th input data, if the predicted value is equal or similar to the corresponding label for each $i = 1, 2, ...$, we can conclude that (6) is zero or near zero (regardless of $g(x)$), which indicates that WES maintains the intrinsic property of the loss function.

- As $g(x)$ is a translated and rescaled function of the $y$-axis symmetry of $f(x)$, a larger value of $\beta$ results in a greater penalty on the region with a lower probability density; that is, WES further constrains the predicted values closer to the label on the region.

- In implementation, $g(x)$ is based on polynomial fitting approximation of $f(x)$ with an appropriate degree; hence, it is infinitely differentiable, as is WES.

- The WES loss function is adaptable. The structure of $g(x)$ changes depending on the distribution of the given label, which means that the function is applicable to all label types.

Figure 1 illustrates $g(x)$ with an example label distribution.

3. DATA MANIPULATION

Three data manipulation stages were carried out using each data type: the generation of label curves, extraction of Fourier signals, and addition of noise.

3.1 Generation of label curves

Four basis curves are generated corresponding to unimodal, skewed unimodal, bimodal, and skewed bimodal distributions, respectively. In the unimodal cases, a basis curve is set to be an inverse function of a cumulative density function (CDF) of $N(0,1)$, for the bimodal cases, a basis curve is comprised as concatenation of inverse CDF of $N(0,1)$ and of $N(0,1/4)$, and for skewed cases, log-normal distribution is used instead (Table 1). By repeating bisymmetrically combined basis pair ($y$-axis symmetry), a sufficiently long curve is generated. Thereafter, we apply uniform normalization to the curve; the maximum and minimum values of the curve are assigned as 1 and 0, respectively, and we refer to the resulting label curve as $L(t)$.

### 3.2 Extraction of Fourier signals

To create appropriate input features corresponding to the label curve, we employ Fourier signals of $L(t)$ as the input data. To be precise, $N$ signals $h_1(t), h_2(t), \ldots, h_N(t)$ are extracted, which correspond to the $N$ absolutely largest coefficients (except for $a_0$) from the $K$th partial sum of Fourier (cosine) series of $L(t)$:

$$L(t) \approx a_0 + \sum_{i=1}^{K} a_i \cos \frac{i\pi t}{M}, \tag{8}$$

where $h_k(t) = \cos \frac{n_k \pi t}{M}$ for $k = 1, 2, \ldots, N$ and $n_k \in \{1, 2, \ldots, K\}$, and $M$ is a number such that $L(t)$ is defind on the interval $[0, M]$.

### 3.3 Addition of noise

A Gaussian noise $\varepsilon(t)$ is added to $h_1(t), h_2(t), \ldots, h_N(t)$ such that $\varepsilon(t) \sim N(0, \sigma^2)$, where $N(0, \sigma^2)$ denotes the Gaussian distribution centered at zero with a standard deviation $\sigma$. We denote the resultant curves as $\widehat{h_1(t)}, \widehat{h_2(t)}, \ldots, \widehat{h_N(t)}$.

## 4. EXPERIMENTS

### 4.1 Overall

A basis curve comprising 2,000 data was generated for each distribution type, and after repeating it 20 times, we constructed a label curve consisting of 40,000 data elements with the values of $M$, $K$, and $N$ in (8) set to 10, 300, and 5, respectively (Figure 2). For each type, the domain of the label curve $L(t)$ was set to $[0, 10]$, and the partial sum of the Fourier series was set to from $i = 1$ to $i = 300$. The $\sigma$ value of the Gaussian noise changed from 0.01 to 0.1 over intervals of 0.01 (10 levels in total), and those were added to the Fourier harmonics corresponding to the five largest coefficients in (8). Different values of $\sigma$ implied varying magnitudes of the Gaussian noise. Furthermore, we varied the hyperparameter $\beta$ in (7) within $\{1.5, 2, 2.5, 3, 4, 5, 6, 7, 8, 9, 10, 15, 20, 25, 30\}$ to test the sensitivity. The loss function $g(x)$ of WES in (7) was approximated by a polynomial of degree 12 for each type of label curve. An experiment batch were constructed for each noise level ($\sigma$), hyperparameter ($\beta$), and loss function, and it consists of 100 ensemble members populated by random initialization.

### 4.2 Loss functions for benchmark

- MSE loss function: The MSE loss function is the most popular second moment error estimator, defined by $\frac{1}{n}\sum_{i=1}^{n}(y_i - \hat{y}_i)^2$, where $y_i$ and $\hat{y}_i$ denote the observed label and predicted value, respectively.

- Mean absolute error (MAE) loss function: The MAE loss function measures the average of the absolute errors defined by $\frac{1}{n}\sum_{i=1}^{n}|y_i - \hat{y}_i|$.

- Huber loss function: The Huber loss function is defined piecewise by $\begin{cases} \frac{1}{2}(y_i - \hat{y}_i)^2 & |y_i - \hat{y}_i| \leq \delta \\ \delta|y_i - \hat{y}_i| - \frac{1}{2}\delta^2 & otherwise \end{cases}$, where $\delta$ is a hyperparameter, which was selected as 0.5, 5, and 10 in this experiment.

- Log-cosh loss function: The log-cosh loss function used in regression analysis is defined by $\frac{1}{n}\sum_{i=1}^{n} \ln(\cosh(y_i - \hat{y}_i))$, where $\cosh(x) = \frac{e^x + e^{-x}}{2}$. The function is approximately equal to $x^2/2$ for small $x$ and to $|x| - \ln 2$ for large $x$.

- Quantile loss function: Whereas MSE estimates the mean of the second moment error, the quantile loss

function focuses on the median of the errors. The function is defined by $\frac{1}{n}\left(\sum_{i=y_i<\hat{y}_i}^{n_1}(\gamma-1)|y_i-\hat{y}_i| + \sum_{i=y_i>\hat{y}_i}^{n_2}(\gamma)|y_i-\hat{y}_i|\right)$, where $n_1 + n_2 = n$ and $\gamma$ is a number to be determined between 0 and 1. If $\gamma = 0.5$, the quantile loss is actually half of the MAE. The hyperparameter $\gamma$ was selected as 0.25 and 0.75 in this study.

4.3 Error metrics

We employed the RMSE and CC, which are typical error metrics in regression analysis, as our basic tools for error measurement. The RMSE measures the quadratic mean of the differences between the label and predicted values, whereas the CC clarifies the statistical relationship between two variables. In addition, we adopted the area of the overlapped sections of two distributions (those of the label and predicted values) as an error metric, since it is a necessary condition for smaller errors. Although a larger common section may not be a necessary and sufficient condition for a better prediction, this research assumed that two overlapping distributions likely lead to a performance gain. Furthermore, the RMSE on the extreme region was employed as another error metric. The extreme region is defined as the region $[0, l_1] \cup [l_2, 1]$ where $P(0 < x < l_1) = P(l_2 < x < 1) = 0.05$, that is, the region corresponds to the rightmost 5% and the leftmost 5% of the probability (0 and 1 in the intervals resulted from the uniform normalization). In this experiment, it was found that $l_1 = 0.200, 0.058, 0.094, 0.039$ and $l_2 = 0.800, 0.540, 0.767, 0.682$ for unimodal, skewed unimodal, bimodal, and skewed bimodal distributed data, respectively. This is an important error metric because it is a key feature of WES to stretch the distribution of predictions by putting greater penalties for the outer values in the distribution. Therefore, a lower RMSE on the region implies a lower error on extreme events, which is a crucial value of WES, in particular, in measures such as financial statistics and flood predictions.

4.4 Structure of neural networks

A total of six layers, consisting of one input layer, three hidden layers, and one output layer, was used in this FFN architecture. The number of nodes in each layer was set to 5, 25, 25, 25, 5, and 1. The constant learning rate, batch size, and epochs were set to 0.01, 512, and 300, respectively. The standard normal distribution and Adam (Kingma & Ba, 2014) were used as the methods of weight initialization and optimization. The linear function as the activation function was located at the output layer, and the sigmoid function defined by $\sigma(x) = \frac{1}{1+e^{-x}}$ was located at the other layers.

5. RESULTS

In this section, we demonstrate the improvements in WES against the other loss functions in terms of mean and best performance. In Figure 3, the performances are intercompared according to the distribution types, error metrics, magnitude of noise (size of $\sigma$), and scaling hyperparameter $\beta$ in Eq. 7. Since we employ the uniform normalization that assigns the minimum and maximum value of the label to be 0 and 1, respectively, the error can be interpreted as a relative error. We evaluated MSE separately since it is the most intuitive and widely used metric. The areas of the common sections between the distributions of the label and the prediction were increased when WES was applied comparing to the other loss function cases; the overlapped area of PDF increased by 3.87% compared to the mean performance of the other loss functions, 0.47% compared to the mean performance of MSE, and 0.8% compared to the best performance of the other loss functions. For RMSE, the performance gains were 0.0044, 0.0016, and 0.0007 (equivalent to 0.44, 0.16 and 0.07%), respectively. For CC, the improvements were 0.0093, 0.0028, and 0.0000, respectively. The improvements for RMSE on the extreme regions (<P5 and >P95) were higher than the evaluation of entire range performance, showing 0.0115, 0.0054, and 0.0027 of RMSE decreases, respectively. A larger overlapping area is only the necessary condition of a lower error. In this research, we hypothesize that increasing the overlapping area would result in overall performance improvements, and the results show that the WES increased the overlapping area, and as a result, RMSE (in particular, in both tail regions) and CC were improved. Moreover, comparing to MSE and the other loss functions, it was noticeable that the performance of the WES is relatively robust to noise, showing less spreads of performance variability against different noise levels. In addition, relatively low performances in the skewed types compared with the non-skewed types were observed for each error metric presumably because of longer tail region of the skewed data.

Figure 4 compares different characteristics between WES and the other loss functions in terms of probability distribution. It is remarkable that WES efficiently stretches the PDF in prediction providing larger penalties on the lower-frequency regions, in contrast to that the saturation boundaries of the other loss functions are further from the label. It was found that the distribution expansion on the left and right 1% ends (i.e., P1 & P99) were 0.0393 & 0.0405, 0.0131 & 0.0537, 0.0175 & 0.0308, and 0.0107 & 0.0438 for unimodal, skewed unimodal, bimodal, and skewed bimodal, respectively.

Mean values of the 100 ensemble predictions in both extreme regions (<P1 and >P99 of distribution) are presented in Table 2. WES could predict the highest values compared to the other loss function on the region regardless of the values of $\sigma$, and the quantile with parameter 0.25 and MSE were comparable. In the other extreme region (Table 2(a)), the MSE and quantile with parameter 0.75 were comparable to WES, however, WES had lowest mean values for large value of $\sigma$. Since the quantile loss function with parameter 0.25 (0.75) is a V-shaped function inclined to the left (right), it gives high penalties on the right (left) extreme of the distribution, therefore the function successfully expanded the right (left) end of the distribution. However, it was found that the function conducted low performance in the left (right) end of distribution because of the low penalties compared to the right end.

It is found that, for a given $\sigma$ and error metric, some loss function was comparable or showed a better performance than WES (e.g. in Figure 3, the gray line that corresponds to MSE error has parts above the red area in the overlapping error metric). However, by the following points of view, we claim that the use of WES, as a strategy for coinciding distributions of the prediction and label, enhanced the prediction performances. First, the other loss functions exhibited their performances highly depended on $\sigma$, whereas WES showed a relatively persistent performance as the noise size $\sigma$ varied. Furthermore, WES was stable, which means that the red spread in Figure 3 was narrow compared to the green spread for all $\sigma$, error metrics, and the label types. Combining above two facts, we conclude that the WES is robust. Since real data inevitably involve noise, WES is an appropriate loss function which is capable of stable prediction against noise. Second, for a given distribution type and $\sigma$, in general, it is always possible to find the hyperparameter $\beta$ that gives better performances than other loss functions for all error metrics (Table S1-16 in Supplementary Information). Furthermore, by the similarity of distributions of hyperparameter $\beta$ in RMSE, CC, and the overlapping region, it could be possible to obtain a $\beta$ which outperforms the other loss functions for multiple error metrics. Third, except for the RMSE on extreme region, the performance was not sensitive to the value of $\beta$, that is, $\beta$ values corresponding to the best performance for different noise levels tend to be distributed within a narrow band (black dots in each graph of Figure 3 are all distributed between $\beta = 1.5$ and $\beta = 5$). Forth, by applying a large value of $\beta$, we may expect a better performance on extreme events, that is, black dots in the fourth column in Figure 3 were distributed with relatively large values compared to the other error metrics. Fifth, it is speculated that the increase in the overlap error metric is closely related to the increase in overlapping region on the both ends of a given distribution type (In the skewed cases, the increase was found only on the lower-frequency region; i.e., the right side in this experiment). Furthermore, regarding to the existence of values larger (smaller) than the maximum (minimum) based on the other loss functions and the improvement in RMSE on extreme region, we claim that WES is an appropriate loss function particularly in the prediction of extreme events.

6.  CONCLUSIONS

Existing error functions exhibit certain drawbacks in terms of two issues: unsatisfactory performances in the extreme regions corresponding to both tail parts of the label distribution (cf. PDF), and instability under input noise. In this study, experiments were conducted to coincide with the distributions of the predicted values and label as an approach to solve the above-mentioned problems. The WES loss function was designed to assign different penalties according to the distribution frequencies; that is, the function assigns a higher penalty in the region with a lower frequency and vice versa. We confirmed by experiments that the WES loss function efficiently increased the area of the common section of the distributions of the predicted values and the label by forcing the predicted values to be close to the label at the tail regions of the distribution. As a result, the errors on the region were diminished, and the overall RMSE and CC were eventually improved. It is anticipated that WES can be applied to time-series predictions, such as rainfall and stock prices, which are highly sensitive to extreme events.

Since the structure of the weighting curve, $g(x)$ depends on the label distribution; it is different from the existing loss functions applied to all label distribution with uniform structure. Such adaptable functioning is an important feature of the WES; it is applicable for all types of label distributions with a different structure. However, a challenge remains to develop a strategy for finding optimal hyperparameter $\beta$ within the learning procedure, although it has a logical problem as $\beta$ is included in the loss function that is the evaluation criteria for measuring errors. In addition to the Gaussian distributed noise, it would be beneficial to test various noise types to further investigate the character of WES.

REFERENCES

Angelova, A., Abu-Mostafa, Y., & Perona, P. (2005). Pruning training sets for learning of object categories. *Proceedings - 2005 IEEE Computer Society Conference on Computer Vision and Pattern Recognition, CVPR 2005*. https://doi.org/10.1109/CVPR.2005.283

Bootkrajang, J., & Kabán, A. (2012). Label-noise robust logistic regression and its applications. *Lecture Notes in Computer Science (Including Subseries Lecture Notes in Artificial Intelligence and Lecture Notes in Bioinformatics)*. https://doi.org/10.1007/978-3-642-33460-3_15

Brodley, C. E., & Friedl, M. A. (1999). Identifying Mislabeled Training Data. *Journal of Artificial Intelligence Research*. https://doi.org/10.1613/jair.606

Charbonnier, P., Blanc-Féraud, L., Aubert, G., & Barlaud, M. (1997). Deterministic edge-preserving regularization in computed imaging. *IEEE Transactions on Image Processing*. https://doi.org/10.1109/83.551699

Cortes, C., & Vapnik, V. (1995). Support-Vector Networks. *Machine Learning*. https://doi.org/10.1023/A:1022627411411

Elman, J. (1990). Finding structure in time* 1. *Cognitive Science*, *211*(1 990), 1–28. Retrieved from http://linkinghub.elsevier.com/retrieve/pii/036402139090002E

Falas, T., & Stafylopatis, A. G. (1999). Impact of the error function selection in neural network-based classifiers. *Proceedings of the International Joint Conference on Neural Networks*, *3*, 1799–1804. https://doi.org/10.1109/ijcnn.1999.832651

Geist, E. L., & Parsons, T. (2008). Distribution of tsunami interevent times. *Geophysical Research Letters*. https://doi.org/10.1029/2007GL032690

Glorot, X., & Bengio, Y. (2010). Understanding the difficulty of training deep feedforward neural networks. *Pmlr*, *9*, 249–256. https://doi.org/10.1.1.207.2059

Glorot, X., Bordes, A., & Bengio, Y. (2011). Deep sparse rectifier neural networks. *Journal of Machine Learning Research*.

Hartley, R., Zisserman, A., Richard, H., Hartley, R., & Zisserman, A. (2003). Multiple View Geometry in Computer Vision, 2nd Edition. *Climate Change 2013 - The Physical Science Basis*. https://doi.org/10.1017/CBO9781107415324.004

Hornik, K., Stinchcombe, M., & White, H. (1989). Multilayer feedforward networks are universal approximators. *Neural Networks*, *2*(5), 359–366. https://doi.org/10.1016/0893-6080(89)90020-8

Huber, P. J. (1964). Robust Estimation of a Location Parameter. *The Annals of Mathematical Statistics*. https://doi.org/10.1214/aoms/1177703732

Jarrett, R. G. (1979). A Note on the Intervals Between Coal-Mining Disasters. *Biometrika*. https://doi.org/10.2307/2335266

Kanamori, T. (2010). Deformation of log-likelihood loss function for multiclass boosting. *Neural Networks*. https://doi.org/10.1016/j.neunet.2010.05.009

Kapur, J. N., & Kesavan, H. K. (1992). *Entropy Optimization Principles and Their Applications*. https://doi.org/10.1007/978-94-011-2430-0_1

Karal, O. (2017). Maximum likelihood optimal and robust Support Vector Regression with lncosh loss function. *Neural Networks*. https://doi.org/10.1016/j.neunet.2017.06.008

Kennedy, B. P., Kawachi, I., Glass, R., & Prothrow-Stith, D. (1998). Income distribution, socioeconomic status,

and self rated health in the United States: Multilevel analysis. *British Medical Journal*. https://doi.org/10.1136/bmj.317.7163.917

Kingma, D. P., & Ba, J. (2014). *Adam: A Method for Stochastic Optimization*. 1–15. https://doi.org/http://doi.acm.org.ezproxy.lib.ucf.edu/10.1145/1830483.1830503

Koenker, R. (2015). Quantile Regression. In *International Encyclopedia of the Social & Behavioral Sciences: Second Edition*. https://doi.org/10.1016/B978-0-08-097086-8.42074-X

Leshno, M., Lin, V. Y., Pinkus, A., & Schocken, S. (1993). Multilayer feedforward networks with a nonpolynomial activation function can approximate any function. *Neural Networks*, *6*(6), 861–867. https://doi.org/10.1016/S0893-6080(05)80131-5

Lin, T. Y., Goyal, P., Girshick, R., He, K., & Dollar, P. (2017). Focal Loss for Dense Object Detection. *Proceedings of the IEEE International Conference on Computer Vision*. https://doi.org/10.1109/ICCV.2017.324

Maas, A. L., Hannun, A. Y., & Ng, A. Y. (2013). Rectifier nonlinearities improve neural network acoustic models. *In ICML Workshop on Deep Learning for Audio, Speech and Language Processing*, *28*.

Murphy, K. P. (2012). Machine learning: a probabilistic perspective (adaptive computation and machine learning series). In *Mit Press. ISBN*.

Reed, S. E., Lee, H., Anguelov, D., Szegedy, C., Erhan, D., & Rabinovich, A. (2015). Training deep neural networks on noisy labels with bootstrapping. *3rd International Conference on Learning Representations, ICLR 2015 - Workshop Track Proceedings*.

Rosasco, L., De Vito, E., Caponnetto, A., Piana, M., & Verri, A. (2004). Are Loss Functions All the Same? *Neural Computation*. https://doi.org/10.1162/089976604773135104

Scalas, E., Gorenflo, R., Luckock, H., Mainardi, F., Mantelli, M., & Raberto, M. (2004). Anomalous waiting times in high-frequency financial data. *Quantitative Finance*. https://doi.org/10.1080/14697680500040413

Schmidhuber, J. (2015). Deep Learning in neural networks: An overview. *Neural Networks*. https://doi.org/10.1016/j.neunet.2014.09.003

Wang, B., & Ho, L. (2002). Rainy season of the Asian-Pacific summer monsson. *Journal of Climate*. https://doi.org/10.1175/1520-0442(2002)015<0386:RSOTAP>2.0.CO;2

Wintz, P. A., & Luecke, E. J. (1969). Performance of Optimum and Suboptimum Synchronizers. *IEEE Transactions on Communication Technology*. https://doi.org/10.1109/TCOM.1969.1090097

Xiao, T., Xia, T., Yang, Y., Huang, C., & Wang, X. (2015). Learning from massive noisy labeled data for image classification. *Proceedings of the IEEE Computer Society Conference on Computer Vision and Pattern Recognition*. https://doi.org/10.1109/CVPR.2015.7298885

Zhu, X., Wu, X., & Chen, Q. (2003). Eliminating Class Noise in Large Datasets. *Proceedings, Twentieth International Conference on Machine Learning*.

# Tables

**Table 1**
**Basis functions for label curves of each distribution type.**

|  | Basis Function |
|---|---|
| Unimodal | Inverse of CDF of $N(0,1)$ |
| Skewed Unimodal | Inverse of CDF of $lnN(0,1)$ |
| Bimodal | Inverse of CDF of $N(0,1)$ and of $N(0,1/4)$ |
| Skewed Bimodal | Inverse of CDF of $lnN(0,1)$ and of $lnN(0,1/4)$ |

**Table 2**
**Performance intercomparison among WES and other loss functions for both tails of each distribution. Std indicates the standard deviation of WES performances based on 15 hyperparameters ($\beta$). Bold font is used to indicate the best performance, and the cell color indicates the best or worst loss function for each experiment.**

| | Left tail of the distribution (P1) ||||||||||||
|---|---|---|---|---|---|---|---|---|---|---|---|---|
| | Unimodal ||| Skewed Unimodal ||| Bimodal ||| Skewed Bimodal |||
| | WES | Others || WES | Others || WES | Others || WES | Others ||
| $\sigma$ | Mean ± Std | Best | Worst | Mean ± Std | Best | Worst | Mean ± Std | Best | Worst | Mean ± Std | Best | Worst |
| 0.01 | 0.062 ± 0.006 | **0.053** | 0.128 | 0.027 ± 0.001 | **0.023** | 0.047 | 0.030 ± 0.001 | **0.029** | 0.070 | 0.013 ± 0.001 | **0.010** | 0.041 |
| 0.02 | 0.062 ± 0.005 | **0.050** | 0.152 | **0.024** ± 0.001 | 0.027 | 0.049 | **0.031** ± 0.002 | 0.033 | 0.072 | 0.014 ± 0.001 | **0.013** | 0.036 |
| 0.03 | 0.058 ± 0.005 | **0.058** | 0.134 | 0.022 ± 0.001 | **0.021** | 0.052 | **0.031** ± 0.001 | 0.034 | 0.064 | 0.013 ± 0.001 | **0.008** | 0.029 |
| 0.04 | 0.059 ± 0.006 | **0.052** | 0.136 | 0.023 ± 0.001 | **0.020** | 0.046 | 0.033 ± 0.002 | **0.032** | 0.064 | 0.013 ± 0.001 | **0.010** | 0.032 |
| 0.05 | 0.060 ± 0.006 | **0.050** | 0.137 | 0.021 ± 0.001 | **0.018** | 0.047 | **0.031** ± 0.002 | 0.034 | 0.068 | **0.012** ± 0.001 | 0.016 | 0.028 |
| 0.06 | 0.062 ± 0.005 | **0.057** | 0.165 | **0.023** ± 0.001 | 0.024 | 0.050 | 0.033 ± 0.002 | **0.031** | 0.070 | **0.012** ± 0.001 | 0.013 | 0.030 |
| 0.07 | **0.060** ± 0.005 | 0.065 | 0.128 | 0.023 ± 0.001 | **0.022** | 0.049 | 0.034 ± 0.002 | **0.031** | 0.073 | 0.013 ± 0.001 | **0.013** | 0.039 |
| 0.08 | **0.060** ± 0.005 | 0.064 | 0.151 | 0.022 ± 0.001 | **0.020** | 0.050 | 0.031 ± 0.001 | **0.029** | 0.067 | 0.013 ± 0.001 | **0.013** | 0.040 |
| 0.09 | 0.059 ± 0.005 | 0.071 | 0.144 | **0.020** ± 0.001 | 0.021 | 0.050 | **0.032** ± 0.001 | 0.033 | 0.072 | **0.012** ± 0.001 | 0.018 | 0.035 |
| 0.10 | **0.064** ± 0.005 | 0.072 | 0.148 | **0.022** ± 0.001 | 0.025 | 0.052 | **0.029** ± 0.001 | 0.031 | 0.069 | 0.014 ± 0.001 | **0.011** | 0.034 |

| | Right tail of the distribution (P99) ||||||||||||
|---|---|---|---|---|---|---|---|---|---|---|---|---|
| | Unimodal ||| Skewed Unimodal ||| Bimodal ||| Skewed Bimodal |||
| | WES | Others || WES | Others || WES | Others || WES | Others ||
| $\sigma$ | Mean ± Std | Best | Worst | Mean ± Std | Best | Worst | Mean ± Std | Best | Worst | Mean ± Std | Best | Worst |
| 0.01 | **0.895** ± 0.005 | 0.892 | 0.824 | 0.761 ± 0.007 | **0.765** | 0.632 | **0.886** ± 0.004 | 0.883 | 0.819 | **0.865** ± 0.005 | 0.860 | 0.760 |
| 0.02 | **0.893** ± 0.004 | 0.886 | 0.820 | 0.758 ± 0.006 | **0.791** | 0.624 | 0.882 ± 0.005 | **0.889** | 0.817 | **0.860** ± 0.004 | 0.858 | 0.762 |
| 0.03 | **0.893** ± 0.004 | 0.890 | 0.817 | **0.758** ± 0.006 | 0.755 | 0.625 | **0.884** ± 0.004 | 0.877 | 0.821 | **0.858** ± 0.004 | 0.849 | 0.759 |
| 0.04 | **0.893** ± 0.005 | 0.886 | 0.823 | **0.757** ± 0.006 | 0.756 | 0.630 | **0.887** ± 0.004 | 0.886 | 0.819 | **0.858** ± 0.004 | 0.852 | 0.756 |
| 0.05 | **0.894** ± 0.005 | 0.883 | 0.815 | **0.769** ± 0.006 | 0.752 | 0.616 | **0.881** ± 0.004 | 0.878 | 0.817 | **0.862** ± 0.004 | 0.859 | 0.749 |
| 0.06 | **0.890** ± 0.004 | 0.878 | 0.818 | **0.758** ± 0.006 | 0.750 | 0.615 | 0.882 ± 0.004 | **0.886** | 0.811 | **0.858** ± 0.004 | 0.849 | 0.749 |
| 0.07 | **0.893** ± 0.004 | 0.880 | 0.803 | **0.759** ± 0.006 | 0.751 | 0.610 | **0.882** ± 0.004 | 0.880 | 0.817 | **0.858** ± 0.004 | 0.843 | 0.747 |
| 0.08 | **0.892** ± 0.005 | 0.876 | 0.807 | **0.758** ± 0.006 | 0.736 | 0.618 | 0.879 ± 0.004 | **0.888** | 0.807 | **0.860** ± 0.004 | 0.848 | 0.750 |
| 0.09 | **0.890** ± 0.005 | 0.870 | 0.778 | 0.753 ± 0.006 | **0.761** | 0.608 | 0.878 ± 0.003 | **0.882** | 0.811 | **0.857** ± 0.004 | 0.851 | 0.715 |
| 0.10 | **0.892** ± 0.004 | 0.876 | 0.794 | **0.758** ± 0.005 | 0.745 | 0.606 | **0.881** ± 0.003 | 0.874 | 0.796 | **0.857** ± 0.004 | 0.839 | 0.743 |

MSE · MAE · Huber 0.5 · Huber5 · Huber10 · Log-Cosh · Quantile 0.25 · Quantile 0.75

# Figures

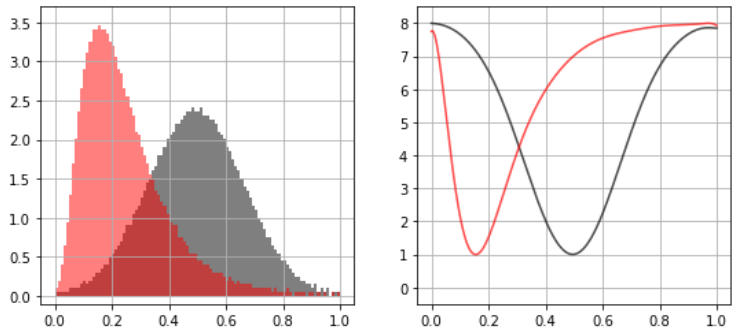

**Figure 1: PDFs of the uniformly normalized labels (left). Corresponding $g(x)$ functions if Eq. 7 generated by polynomial approximation ($\beta = 8$) (right). Black and red colors indicate unimodal and skewed unimodal, respectively.**

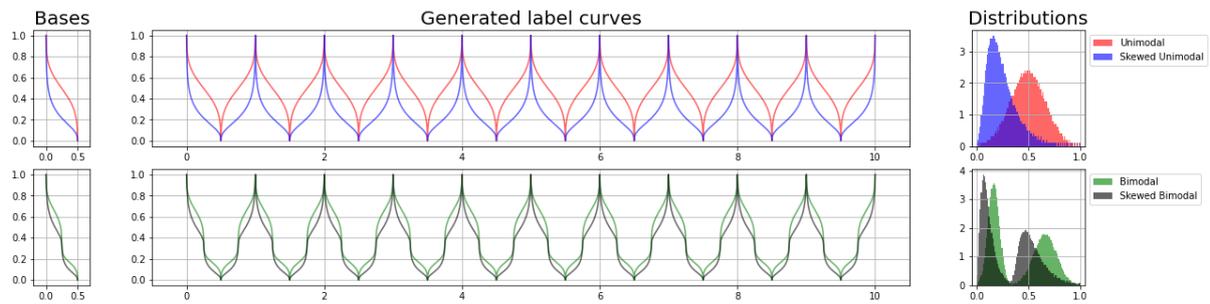

**Figure 2: Basis curves (left) and PDF (right) of four distributions in Table 1. Label series of bisymmetrically combined basis pair (center). Red, blue, green and black indicate unimodal, skewed unimodal, bimodal, and skewed bimodal distributions, respectively.**

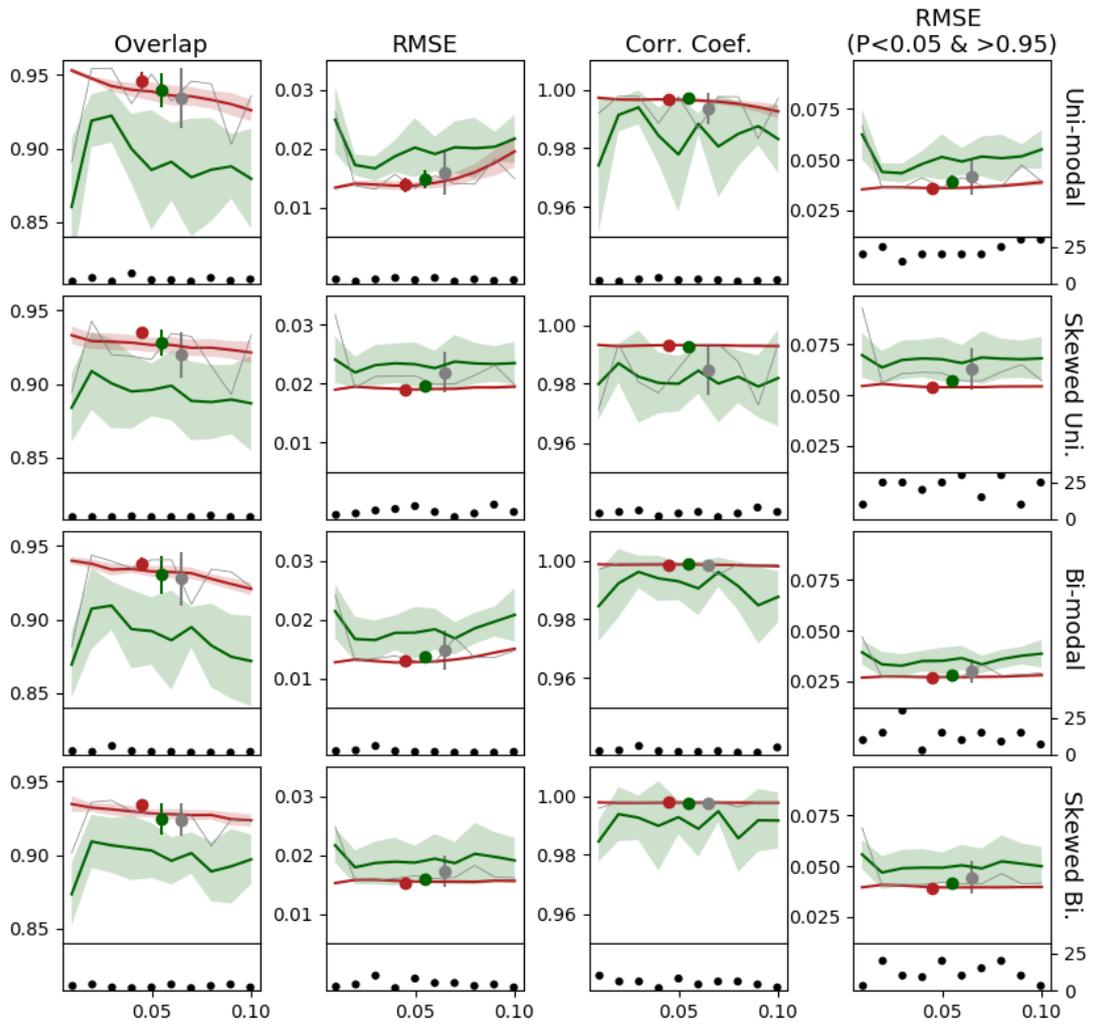

Figure 3: **Performances of loss functions against different noise levels ($\sigma$) (x-axis) for error metrics (overlapped area ratio, RMSE, CC, and RMSE in tail regions) (columns) and label distributions (uni-modal, skewed uni-modal, bi-modal, and skewed bi-modal) (rows). Y-axis of upper-panel and lower-panel in each subplot indicate error metric and $\beta$, respectively. Solid lines represent mean of WES for different $\beta$ (red), MSE (gray), and the mean of the loss functions in Table 2 (green), and shades represent their spreads (one standard deviation). Closed circles and error bars indicate the best performance and spread among $\sigma$, respectively. Black dots show variability of $\beta$ associated with the best performance through varying noise levels.**

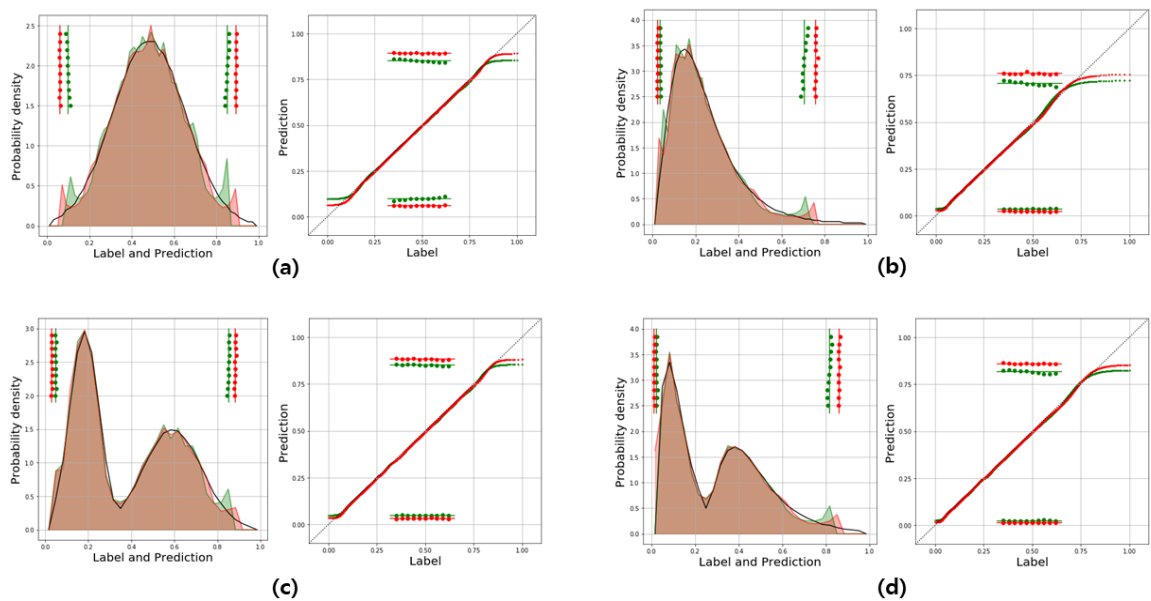

Figure 4: Performance comparisons in the distribution (left-panel) and scatter plot (right-panel) corresponding to unimodal (a), skewed unimodal (b), bimodal (c) and skewed bimodal (d). Red, green, and black represent the averaged results on WES and the loss functions in Table 2, and the label, respectively. Dots represent the mean of 1% both tail values corresponding to each $\sigma$ from 0.01 to 0.1 (from top to bottom in the distribution, from left to right in the scatter plot).